\documentclass{article}
\usepackage{ijcai22}

\usepackage{times}
\usepackage{soul}
\usepackage{url}
\usepackage[hidelinks]{hyperref}
\usepackage[utf8]{inputenc}
\usepackage[small]{caption}
\usepackage{graphicx}
\usepackage{amsmath}
\usepackage{amssymb}
\usepackage{amsthm}
\usepackage{booktabs}
\usepackage{algorithm}
\usepackage{algorithmic}
\usepackage{multirow}
\usepackage{multicol}
\usepackage{color}
\title{Entity-aware and Motion-aware Transformers for Language-driven Action Localization in Videos}

\author{
Shuo Yang
\and
Xinxiao Wu\thanks{corresponding author}
\affiliations
Beijing Laboratory of Intelligent Information Technology,\\
School of Computer Science, Beijing Institute of Technology\\
\emails
\{shuoyang,wuxinxiao\}@bit.edu.cn
}

\begin{document}

\maketitle

\begin{abstract}
Language-driven action localization in videos is a challenging task that involves not only visual-linguistic matching but also action boundary prediction. 
Recent progress has been achieved through aligning language query to video segments, but estimating precise boundaries is still under-explored. In this paper, we propose entity-aware and motion-aware Transformers that progressively localizes actions in videos by first coarsely locating clips with entity queries and then finely predicting exact boundaries in a shrunken temporal region with motion queries.
The entity-aware Transformer incorporates the textual entities into visual representation learning via cross-modal and cross-frame attentions to facilitate attending action-related video clips. %
The motion-aware Transformer captures fine-grained motion changes at multiple temporal scales via integrating long short-term memory into the self-attention module to further improve the precision of action boundary prediction.
Extensive experiments on the Charades-STA and TACoS datasets demonstrate that our method achieves better performance than existing methods. Codes are available at \hyperlink{https://github.com/shuoyang129/EAMAT}{https://github.com/shuoyang129/EAMAT}. %

\end{abstract}

\section{Introduction}
Language-driven action localization, also called temporal video grounding or video moment retrieval, aims to localize the start and end frames of an action relevant to the language query. It has attracted growing attention for its wide applications, such as robotic navigation and video understanding.
This task is challenging since it requires not only aligning the  language query to video segments but also estimating the temporal boundaries of the desired action. %

\begin{figure}
    \centering
    \includegraphics[width=\linewidth]{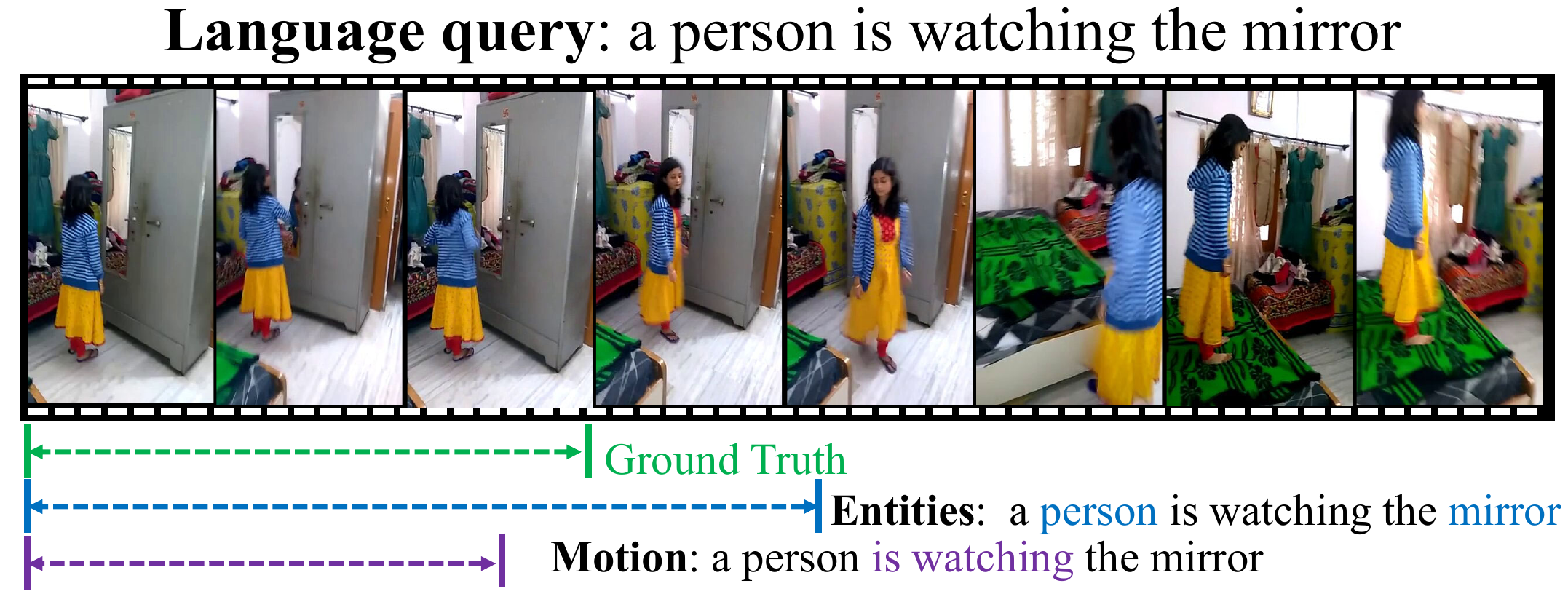}
    \caption{Illustration of coarse-to-fine action localization using entity and motion queries.}
    \label{fig:intro}
\end{figure}
\begin{figure*}
    \centering
    \includegraphics[width=0.98\textwidth]{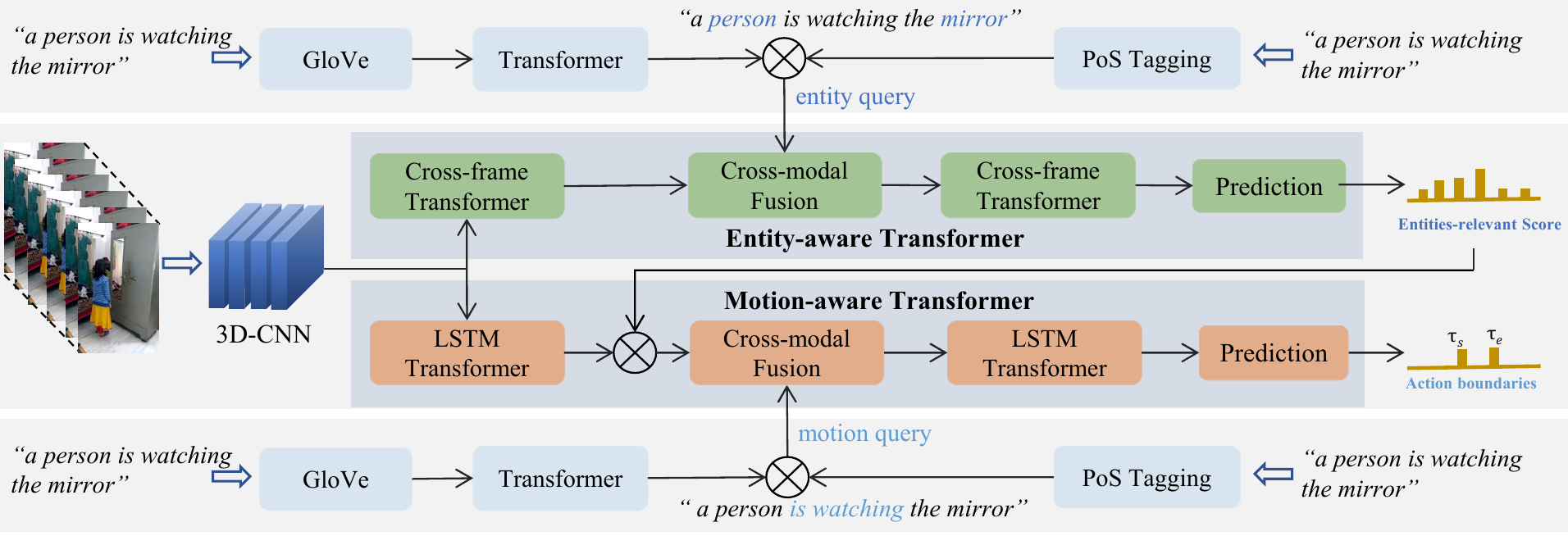}
    \caption{Overview of our entity-aware and motion-aware Transformers for language-driven action Localization.}
    \label{fig:pipline}
\end{figure*}

Tremendous effects have been devoted to the alignment between language query and video segments. Several early studies~\cite{hendricks2017localizing} resort to learning a common visual-textual embedding space by pushing dissimilar or pulling similar visual features and linguistic features. 
Later, in order to explore more detailed semantics for visual-textual alignment, some methods~\cite{chen2019semantic} extract semantic concepts of actions or objects to enrich the holistic features of both video and language. 
In more recent years, various attention operations have been proposed to learn elaborate cross-modal relations, such as self-modal or cross-modal graph attention~\cite{liu2020jointly}, context-query attention~\cite{zhang2021natural} and local-global interaction~\cite{mun2020local}. 
All these methods mainly focus on learning and aligning the visual and linguistic representations for language-driven action localization without considering the explicit modeling of finer action boundaries for precise localization.

This paper investigates a coarse-to-fine strategy to progressively estimate the action boundaries in untrimmed videos with high precision. With this in mind, we propose entity-aware and motion-aware Transformers that first coarsely locate video clips from the entire video with textual entities and then finely predict exact boundaries in a shrunken temporal region with motion  queries. For example, as illustrated in Figure 1, the query sentence of ``\textit{a person is watching the mirror}" can be divided into two types of information: the entities of ``\textit{person \& mirror}" and the motion of ``\textit{is watching}". Our method first finds the frames in which the ``\textit{person \& mirror}" appear, and then localizes the start and end boundaries between which the ``\textit{is watching}" happens.

To be more specific, the entity-aware Transformer incorporates textual entities of language query into visual representation learning via cross-modal attention. The learned visual features are capable of attending to the salient action-related objects so as to facilitate selecting action-related video clips.  
Moreover, cross-frame attention is employed to leverage contextual information from adjacent frames to learn more robust entity features.
Starting with more accessible entities, the entity-aware Transformer narrows the searching space for action localization by coarsely locating temporal regions where the desired action is more likely to happen.

An action  consists of sequential motions and large motion changes usually lie on the action boundaries. To improve the action localization precision, it is significantly essential to capture the fine-grained motion changes in videos. So we propose a motion-aware Transformer that integrates a long short-term memory cell into the self-attention module in Transformer.
Intuitively, the long short-term memory cell is a natural way to capture the consecutive local motion changes, and we apply it at multiple temporal scales to deal with various durations of the same action.
Transformer is capable of modeling the long-range dependency and has been proved its effectiveness in many visual and linguistic tasks~\cite{vaswani2017attention}, and it is reasonable to use it for modeling the global motion interactions. Therefore, our motion-aware Transformer can capture fine-grained motion changes at multiple time granularities, which benefits a lot to localizing the exact boundaries of desired actions.

The main contributions of this paper are summarized as follows: 
(1) We propose a coarse-to-fine framework for language-driven  action  localization, which extracts detailed entity and motion queries to  progressively  estimate  the  action  boundaries with high precision.
(2) We propose entity-aware and motion-aware Transformers as an effective implement of the coarse-to-fine localization, where the newly designed motion-aware Transformer models fine-grained motion changes at multiple temporal scales by integrating long short-term memory into self-attention.
(3) Extensive experiments on popular benchmarks, Charades-STA and TACoS, demonstrate that the proposed method performs favorably against existing methods.

\section{Related Work}
The language-driven action localization task is firstly proposed in ~\cite{gao2017tall,hendricks2017localizing}. It is tackled by first generating proposals with manually designed temporal bounding boxes %
and then ranking the proposals by the given language query. %
To enhance the visual and linguistic representations, ACRN ~\cite{liu2018attentive} proposes a memory attention mechanism to emphasize the language-related visual features with context information. SCDM~\cite{yuan2020semantic} modulates the temporal convolution operations for better correlating and composing the sentence related video contents.
2D-TAN~\cite{zhang2020learning} uses a 2D temporal adjacent network to learn contextual and structural information between adjacent moment candidates. 
MAST~\cite{zhang2021multi} aggregates multi-stage features to represent moment proposals using a BERT-variant Transformer backbone.
These proposal-based methods are relatively inefficient since a large number of proposals causes redundant computation. Moreover, the boundaries of  proposals are fixed, leading to inflexible estimations.

To mitigate the defects of manually designed proposals, proposal-free methods~\cite{zeng2020dense,yuan2019to,hahn2019tripping,lu2019debug,wu2020tree,li2021proposal,zhao2021cascaded} are proposed to directly predict the action boundaries through visual and linguistic representation alignment.
ExCL~\cite{ghosh2019excl} and SeqPAN~\cite{zhang2021parallel} predict the start and end time by leveraging the cross-modal interaction between the text and video;
LGI~\cite{mun2020local}, CSMGAN~\cite{liu2020jointly}, FIAN ~\cite{qu2020fine}, CBLN~\cite{liu2021context}, SMIN~\cite{wang2021structured}, I$^2$N~\cite{ning2021interaction} explore the local and global context information for accurate localization. 

Rather than mainly focusing on aligning the visual and linguistic representations in the aforementioned methods,
we attempt to achieve high localization precision by designing a progressive strategy that first narrows the target regions and then localizes the finer boundaries. 
The most related work to our method is VSLNet~\cite{zhang2021natural} that searches for the target action within a highlighted region, which extends the target action segment by a simple hyper-parameter in a span-based question answering framework.
In contrast, our method coarsely locates the temporal region first by the apparent action-related entities, and then finely predicts the action boundaries by explicitly modeling fine-grained motion changes at both short and long times, achieving high precision, good generalization and interpretability.

\section{Our Method}
Given an untrimmed video $V=\{v_t\}_{t=1}^{T}$ and a language query $S=\{w_i\}_{i=1}^{N}$ where $v_t$ represents the $t$-th video frame, $w_i$ represents the $i$-th word, and $T$ and $N$ represent the number of video frames and text words, respectively, our task aims to localize the target action boundaries $(\tau_s,\tau_e)$ where $\tau_s$ and $\tau_e$ represent the start and end frames of the action corresponding to the query, respectively. 
As shown in Figure \ref{fig:pipline}, our method has two main components: an entity-aware Transformer and a motion-aware Transformer. The former incorporates the entity terms, \textit{i.e.,} subjects and objects, of the language query into the visual representation learning to filter out the video clips that have no action-relevant entities. %
The latter captures fine-grained motion changes by integrating a long short-term memory cell into the self-attention module guided by the motion terms, \textit{i.e.,} verbs, of the language query to refine the start and end frames.

\subsection{Entity and Motion Query Extraction}
\label{sec:language}
We encode the input language query $S$ into entity query features $\textbf{F}_{q}^{e}$ and motion query features $\textbf{F}_{q}^{m}$. The words in $S$ are classified into three classes: entity, motion, and others, by using the part of speech tags\footnote{https://www.nltk.org/} of the words. 
The classification probabilities of the word $i$ are denoted by ${\textbf{p}}_i = [p_i^{e}; p_i^{m}; p_i^{o}] \in \{0,1\}^3$. 
For the word $i$, if its part of speech tag is related to entity  (\textit{i.e.}, noun, adjective), then $\textbf{p}_i=[1,0,0]$; if its part of speech tag  is related to motion (\textit{i.e.},verb, adverb), then $\textbf{p}_i=[0,1,0]$; otherwise, $\textbf{p}_i=[0,0,1]$.

The word features $\textbf{Q}=[\textbf{w}_1,\textbf{w}_2,\cdots,\textbf{w}_N]^{\top} \in \mathbb{R}^{N\times d_w}$ of $S$ are first initialized using the GloVe embedding ~\cite{pennington2014glove}, where $\textbf{w}_i$ denotes the $i$-th word feature with dimension $d_w$ and $N$ denotes the word number in the language query. And then a Transformer block is used to learn the relationships between the words, given by
\begin{equation}\small
    \textbf{F}_{q} = Transformer_q( FC_1(\textbf{Q}) )
\end{equation}
where $\textbf{F}_{q}=[\textbf{f}_{q,1},\textbf{f}_{q,2}, \cdots,\textbf{f}_{q,N}]^{\top} \in \mathbb{R}^{N\times d}$ are the learned linguistic query features; $FC_1(\cdot)$ is a fully connected layer that projects the word feature from dimension $d_w$ to $d$; $Transformer_q(\cdot)$ is a standard Transformer block, as shown in Figure \ref{fig:transformer}(a), which consists of multi-head self-attention, residual connection, layer normalization and feed-forward network.
Finally, the entity query features $\textbf{F}_{q}^{e}=[\textbf{f}_{q,1}^{e},\textbf{f}_{q,2}^{e}, \cdots,\textbf{f}_{q,N}^{e}]^{\top} \in \mathbb{R}^{N\times d}$ and motion query features $\textbf{F}_{q}^{m}=[\textbf{f}_{q,1}^{m},\textbf{f}_{q,2}^{m}, \cdots,\textbf{f}_{q,N}^{m}]^{\top} \in \mathbb{R}^{N\times d}$ of the language query  are calculated by
\begin{equation}\small
\label{equ:w_entitiy}  
  \begin{array}{l}
    \textbf{f}_{q,i}^{e} =\textbf{f}_{q,i} \cdot (p_i^{e} + p_i^{o}),\quad
    \textbf{f}_{q,i}^{m} =\textbf{f}_{q,i} \cdot (p_i^{m} + p_i^{o}).
  \end{array}
\end{equation}

\subsection{Entity-aware Transformer} \label{sec:entities_perception}
As the objects in video are more easily accessible and provide rich indication information for actions, it is natural to narrow down the searching space from all the frames to the actual relevance using the entities in the language query.
So we propose an entity-aware Transformer to coarsely select the video clips that are related to the input entity queries.
As illustrated in Figure \ref{fig:pipline}, the entity-aware Transformer first learns relationships between video frames via cross-frame attention to provide more contextual information, then fuses the entity query features into each video frame via cross-modal attention, next, attends complementary information across different frames via cross-frame attention, and finally predicts an action-relevant score for each frame to indicate whether the frame is action-relevant or not.
According to the predicted action-relevant scores, we select action-relevant video clips where the desired action may happen.
\begin{figure}
    \centering
    \includegraphics[width=\linewidth]{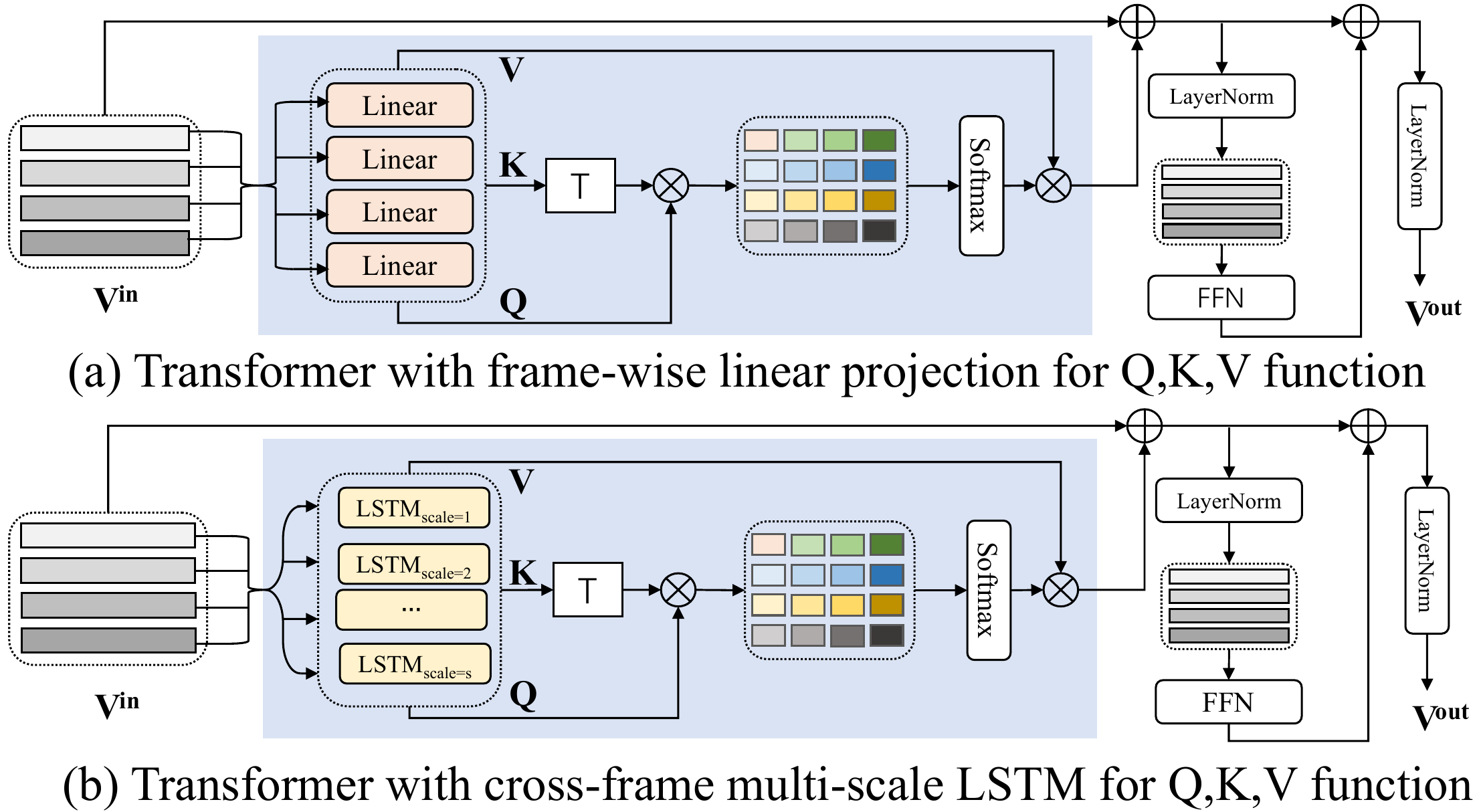}
    \caption{Standard Transformer (a) and our LSTM Transformer (b).}
    \label{fig:transformer}
\end{figure}

\noindent\textbf{Cross-frame Transformer.} %
For each video $V$, we extract its visual features $\textbf{F}_{v}=[\textbf{f}_{v,1},\textbf{f}_{v,2},\cdots,\textbf{f}_{v,T}]^{\top} \in \mathbb{R}^{T \times d_v}$ by a pre-trained 3D ConvNet, where $\textbf{f}_{v,i}$ denotes the $i$-th visual feature with dimension $d_v$ that is computed on a short video clip  and $T$ denotes the number of features.  
A standard Transformer block is then used to attend contextual information across different frames:
\begin{equation}\small
\label{equ:transformer_e}
    \textbf{F}_{v}^{e} = Transformer_e(FC_2(\textbf{F}_{v}))
\end{equation}
where $\textbf{F}_{v}^{e}=[\textbf{f}_{v,1}^{e},\textbf{f}_{v,2}^{e}, \cdots,\textbf{f}_{v,T}^{e}]^{\top} \in \mathbb{R}^{T\times d}$ are the updated visual features that pay more attention to the entities; $FC_2(\cdot)$ is a fully connected layer that projects the visual feature from dimension $d_v$ to $d$; 
 $Transformer_e(\cdot)$ represents a standard Transformer block, as shown in Figure \ref{fig:transformer}(a). %
We concentrate on the appearance information of frames without considering the temporal information between them, so no position embedding is input to the Transformer.

\noindent\textbf{Cross-modal Fusion.}
We introduce the context-query attention (CQA)~\cite{zhang2021natural} to integrate the entity query features into visual features of each frame. 
Given the visual features $\textbf{F}_{v}^{e}$ and the entity query features $\textbf{F}_{q}^{e}$, CQA first computes the their similarity $\textbf{S} = \textbf{F}_{v}^{e}\cdot {\textbf{F}_{q}^{e}}^{\top} \in \mathbb{R}^{T\times N}$,  followed by a row-wise and column-wise softmax normalization to obtain two similarity matrices $\mathcal{\textbf{S}}_r$ and $\mathcal{\textbf{S}}_c$. Then two attention weights are derived by $\mathcal{A}_{VQ}=\mathcal{\textbf{S}}_r\cdot \textbf{F}_{q}^{e}$ and $\mathcal{A}_{QV}=\mathcal{\textbf{S}}_r\cdot \mathcal{\textbf{S}}_c^{\top}\cdot {\textbf{F}_{v}^{e}}$. 
The entity-aware visual features  $\textbf{F}^{ve}$ are computed by
\begin{equation}\small
    \textbf{F}^{ve} = FC_3([\textbf{F}_v^{e};\mathcal{A}_{VQ};{\textbf{F}_v^{e}}\odot \mathcal{A}_{VQ};{\textbf{F}_v^{e}}\odot \mathcal{A}_{QV}])
\end{equation}
where $\textbf{F}^{ve}=[\textbf{f}_1^{ve},\textbf{f}_2^{ve},\cdots,\textbf{f}_T^{ve}]^{\top} \in \mathbb{R}^{T\times d}$; $\odot$ denotes element-wise multiplication; [$\cdot$] is concatenation; $FC_3(\cdot)$ is a fully connected layer that projects the concatenated feature from dimension $4d$ to $d$.

\noindent\textbf{Prediction.}
 We calculate the action-relevant score $\textbf{P}_e=[p_{e,1},p_{e,2},\cdots, p_{e,T}]^{\top} \in \mathbb{R}^{T}$ of video frames using two fully connected layers for action location prediction:
\begin{equation}\small
\label{equ:action_relavant_score}
    \textbf{P}_e = sigmoid(FC_5(ReLU(FC_4(\textbf{F}^{ve}))))
\end{equation}
where the output feature dimensions of $FC_4(\cdot)$ and  $FC_5(\cdot)$ are $\frac{d}{2}$ and 1, respectively.
The higher the action-relevant score is, the higher the probability that the corresponding frame is selected as  action regions.

\subsection{Motion-aware Transformer} \label{sec:motion_perception}

Given the coarsely located video clips by the entity-aware Transformer, we propose a motion-aware Transformer to refine the action boundaries by capturing both fine-grained local and global motion changes. As shown in Figure \ref{fig:pipline}, the motion-aware Transformer first learns contextual motion information by a novel LSTM Transformer, then attends the action-relevant parts by the action-relevant score of entity-aware Transformer and fuses motion query features into them via cross-modal attention, next, captures the fine-grained motion changes and global motion changes via the LSTM Transformer, and finally predicts the action boundaries.

\noindent \textbf{LSTM Transformer.} 
The standard Transformer can capture global motion changes due to its capability of modeling long-range dependency where the self-attention module plays a vital role. The self-attention module first conducts linear projections on each input unit to obtain query, key, and value features, and then uses the similarity of query-key feature to aggregate the value features, as shown in Figure \ref{fig:transformer}(a). %
However, the linear projections cannot capture local motion changes in successive frames. Thus we replace the linear projection with a LSTM cell, as shown in Figure \ref{fig:transformer}(b), which learns sequential local motion changes in videos. In order to deal with the duration variations of the same action in different videos, we apply the long short-term memory at multiple temporal scales.

Specifically, given an input sequence $\textbf{V}^{in}=\{\textbf{v}_i^{in}\}_{i=1}^{T}$, the multi-head self-attention module of our LSTM Transformer is given by $MSA(\textbf{f}_Q,\textbf{f}_K,\textbf{f}_V)=[h_1,h_2,\cdots,h_n]$ where a single head is calculated as $h_i=SA_i(\textbf{f}_Q,\textbf{f}_K,\textbf{f}_V) = softmax(\textbf{f}_Q \textbf{f}_K^\top/\sqrt{d})\textbf{f}_V$, where $d$ is the dimension of intermediate features and $\textbf{f}_\nu= LSTM_\nu^{S}(\textbf{V}^{in}), \nu \in [Q,K,V]$.
The $LSTM^{S}$ is a multi-scale version of the LSTM, denoted as $LSTM^{S}(\textbf{V}^{in}) = [L_1(\textbf{V}^{in});L_2(\textbf{V}^{in});\cdots;L_S(\textbf{V}^{in})]$, where $[\cdot]$ is concatenation. The $s$-th scale LSTM is calculated by
\begin{equation}
\small
    L_s(\textbf{V}^{in})=LSTM^s(\cdots,\textbf{V}_{i-2s}^{in}, \textbf{V}_{i-s}^{in}, \textbf{V}_{i}^{in}, \textbf{V}_{i+s}^{in},\textbf{V}_{i+2s}^{in}, \cdots).
\end{equation}
One-time running of $L_s$ can update input sequence every $s$ frames, and sliding $L_s$ in the input sequence one frame for $s$ times can update all input sequence.

For each video $V$ and its visual features $\textbf{F}_{v}=[\textbf{f}_{v,1},\textbf{f}_{v,2},\cdots,\textbf{f}_{v,T}]^{\top} \in \mathbb{R}^{T \times d_v}$, the LSTM Transformer is used to learn both fine-grained local and global motion changes:
\begin{equation}\small
\label{equ:transformer_m}
    \textbf{F}_v^{m} = Transformer_m(FC_2(\textbf{F}_v))
\end{equation}
where $\textbf{F}_{v}^{m}=[\textbf{f}_{v1}^{m},\textbf{f}_{v2}^{m}, \cdots,\textbf{f}_{vN}^{m}] \in \mathbb{R}^{T\times d}$ are the updated motion features that pay more attention to the motion changes;  $FC_2(\cdot)$ is the fully connected layer used in Equation (\ref{equ:transformer_e}) that projects the visual feature from dimension $d_v$ to $d$;
 $Transformer_m(\cdot)$ represents the LSTM Transformer.

\noindent \textbf{Cross-modal Fusion.}
Before cross-modal fusion, we first attend the action-relevant video frames by the action-relevant score $\textbf{P}_e$ %
in a soft manner: $\textbf{F}_v^{m} = \textbf{P}_e \odot \textbf{F}_v^{m}$, where $\odot$ is an element-wise multiplication.
Then motion query features $\textbf{F}_{q}^{m}$ (calculated in Section \ref{sec:language}) are fused into the visual motion representation $\textbf{F}_{v}^{m}$ via CQA (described in Section \ref{sec:entities_perception}) to obtain the motion-aware visual features:
\begin{equation}\small
    \textbf{F}^{vm} = CQA(\textbf{F}_q^{m}, \textbf{F}_v^{m})
\end{equation}
where $\textbf{F}^{vm}=[\textbf{f}_1^{vm},\textbf{f}_2^{vm},\cdots,\textbf{f}_T^{vm}]^{\top} \in \mathbb{R}^{T\times d}$.

\noindent \textbf{Prediction.}
The start scores $\textbf{S}_s \in \mathbb{R}^{T}$ and the end scores $\textbf{S}_e \in \mathbb{R}^{T}$ for target action segment are predicted  by a two-branch network consisting of two fully connected layers:
\begin{equation}\small
    \begin{array}{l}
        \textbf{S}_s = FC_7(ReLU(FC_6(\textbf{F}^{vm}))) \\
        \textbf{S}_e = FC_9(ReLU(FC_8(\textbf{F}^{vm})))
  \end{array}
\end{equation}
where the output feature dimensions of $FC_i(\cdot), i\in\{6,8\}$ and $FC_j(\cdot), j\in\{7,9\}$ are $\frac{d}{2}$ and 1, respectively.
Then the probability distributions of action start and end boundaries are computed by $\textbf{P}_s^{b} = softmax(\textbf{S}_s), \textbf{P}_e^{b} = softmax(\textbf{S}_e) \in \mathbb{R}^{T}$. 
Finally, the predicted start and end boundaries of target action segment are derived by maximizing the joint probability:
\begin{equation}\small
    \begin{array}{c} 
      (\hat{\tau_s},\hat{\tau_e}) = \arg\max_{t_s,t_e}\textbf{P}_s^{b}(t_s) \times \textbf{P}_e^{b}(t_e),\vspace{1ex} \\ 
      p_{se}^{b} = \textbf{P}_s^{b}(\hat{\tau_s}) \times \textbf{P}_e^{b}(\hat{\tau_e})
    \end{array}
\end{equation}
where $p_{se}^{b}$ is the optimized score of the predicted boundaries $(\hat{\tau_s},\hat{\tau_e})$. 
We also apply another branch of two fully connected layers network to predict a inner probability for each frame as a auxiliary task only for training~\cite{wang2021structured}. Let $\textbf{P}^{in}=[\textbf{p}^{in}_{1},\textbf{p}^{in}_{2},\cdots,\textbf{p}^{in}_{T}]^{\top} \in \mathbb{R}^{T}$ denote the probability of being action frames, calculated by %
\begin{equation}\small
    \textbf{P}^{in} = sigmoid(FC_b(ReLU(FC_a(\textbf{F}^{vm}))))
\end{equation}
where the output feature dimensions of $FC_a(\cdot)$ and $FC_b(\cdot)$ are $\frac{d}{2}$ and 1, respectively.

\subsection{Training Objective}
Given the predicted probability distribution of action boundaries $\textbf{P}_s^{b}$ and $\textbf{P}_e^{b}$ , the training objective for action boundary prediction is formulated by
\begin{equation}\small
    \mathcal{L}^{boundary} = f_{XE}(\textbf{P}_{s}^{b},\tau_s) + f_{XE}(\textbf{P}_{e}^{b},\tau_e)
\end{equation}
where $f_{XE}(\cdot)$ is a cross-entropy function, and $(\tau_s,\tau_e)$ are the ground-truth boundaries.
Given the inner probability $\textbf{P}^{in}$, the training objective for action frame prediction is formulated by
\begin{equation}\small
    \mathcal{L}^{inner} = f_{BXE}(\textbf{P}^{in},\textbf{Y}^{in})
\end{equation}
where $f_{BXE}(\cdot)$ is a binary cross-entropy function and $\textbf{Y}^{in}=\{y^{in}_{i}\}_{i=1}^{T} \in \{0,1\}$, when $\tau_s \leq i \leq \tau_e$, $y^{in}_{i}=1$, otherwise $y^{in}_{i}=0$.
The overall  objective  is given by
\begin{equation}\small
\label{equ:loss}
     \mathcal{L} = \lambda_1 \mathcal{L}^{boundary} + \lambda_2\mathcal{L}^{inner}
\end{equation}
where $\lambda_1$ and $\lambda_2$ are hyper-parameters.

\section{Experiments}
\label{sec:experiments}
\subsection{Datasets and Evaluation Metrics}
We evaluate our method on two datasets: \textbf{Charades-STA}~\cite{gao2017tall} and \noindent \textbf{TACoS}~\cite{regneri2013grounding}. The Charades-STA dataset is built on the Charades dataset~\cite{sigurdsson2016hollywood} and contains 16,128 annotations, including 12,408 for training and 3,720 for test. %
The TACoS dataset is built on the MPII Cooking Compositive dataset~\cite{rohrbach2012script} and contains 18,818 annotations, including 10146 for training, 4589 for validation, and 4083 for test. 

We use the metrics of $R@n; IoU=\mu$ and $mIoU$ for evaluation. $R@n; IoU=\mu$ denotes the percentage of test samples that have at least one result whose IoU with ground-truth is larger than $\mu$ in top-n predictions and $mIoU$ denotes the average IoU over all test samples. 
We set $n=1$ and $\mu \in [0.3, 0.5, 0.7]$.

\subsection{Implementation Details}
Following the previous methods, 3D convolutional features (C3D for TACoS, and I3D for Charades-STA) are extracted to encode videos.
We adopt Adam~\cite{kingma2014adam} for optimization with an initial learning rate of 5e-4 and a linear decay schedule.
The loss weights  $\lambda_1$ and $\lambda_2$ in Equation (\ref{equ:loss}) are set to 1 and 10, respectively.
The number of Transformer Blocks is set to 1 and 3 for early and late Transformers in entity-aware and motion-aware Transformers. The feature dimension of all intermediate layers is set to 512, the head number of multi-head self-attention is set to 8, the layer number and scale number of long short-term memory are set to 1 and 3, respectively.

\subsection{Comparison Results}
\begin{table}
\small
\centering
\begin{tabular}{l|ccc|c}
\hline
\hline
\multicolumn{1}{c|}{\multirow{2}{*}{Methods}}  & \multicolumn{3}{c|}{$R@1; IoU = \mu$} & \multirow{2}{*}{$mIoU$} \\
            & \multicolumn{1}{c}{0.3}  & \multicolumn{1}{c}{0.5}  & \multicolumn{1}{c|}{0.7} \\ \hline
VSLNet~\cite{zhang2021natural}     & 70.46 & 54.19 & 35.22 & 50.02 \\
LGI~\cite{mun2020local}       & 72.96 & 59.46 & 35.48 & 51.38 \\
DeNet~\cite{zhou2021embracing}  & - & 59.7 & 38.52 & - \\
SS~\cite{ding2021support} & - & 60.75  & 36.19 & - \\
CPNet~\cite{li2021proposal} & 71.94  & 60.27  & 38.74 & 52.00  \\
ACRM~\cite{tang2020frame}    & 73.47 & 57.53 & 38.33 & -  \\ 
ICG~\cite{nan2021interventional}    & 67.63 & 50.24 & 32.88 & 48.02  \\
CPN~\cite{zhao2021cascaded}    & 68.48 & 51.07 & 31.54 & 48.08  \\
SeqPAN~\cite{zhang2021parallel} &  73.84 & 60.86 & 41.34 & 53.92  \\
CBLN~\cite{liu2021context}   & - & 61.13 & 38.22 & -\\ \hline
\multicolumn{1}{c|}{Ours}    & \textbf{74.19} & \textbf{61.69} & \textbf{41.96} & \textbf{54.45}  \\
\hline \hline
\end{tabular}
\caption{Comparison with the state-of-the-art methods on the Charades-STA dataset.}
\label{tab:sota_charades}
\end{table}

We compare our method with the latest state-of-the-art methods on the Charades-STA and TACoS datasets in  
Table \ref{tab:sota_charades} and Table \ref{tab:sota_tacos}, respectively. 
From the results, it is interesting to observe that our method achieves the best performance in terms of all evaluation metrics on both two datasets, clearly validating the superiority of the proposed entity-aware and motion-aware Transformers on improving the localization precision via a coarse-to-fine strategy.

\begin{figure*}
    \centering
    \includegraphics[width=0.9\linewidth]{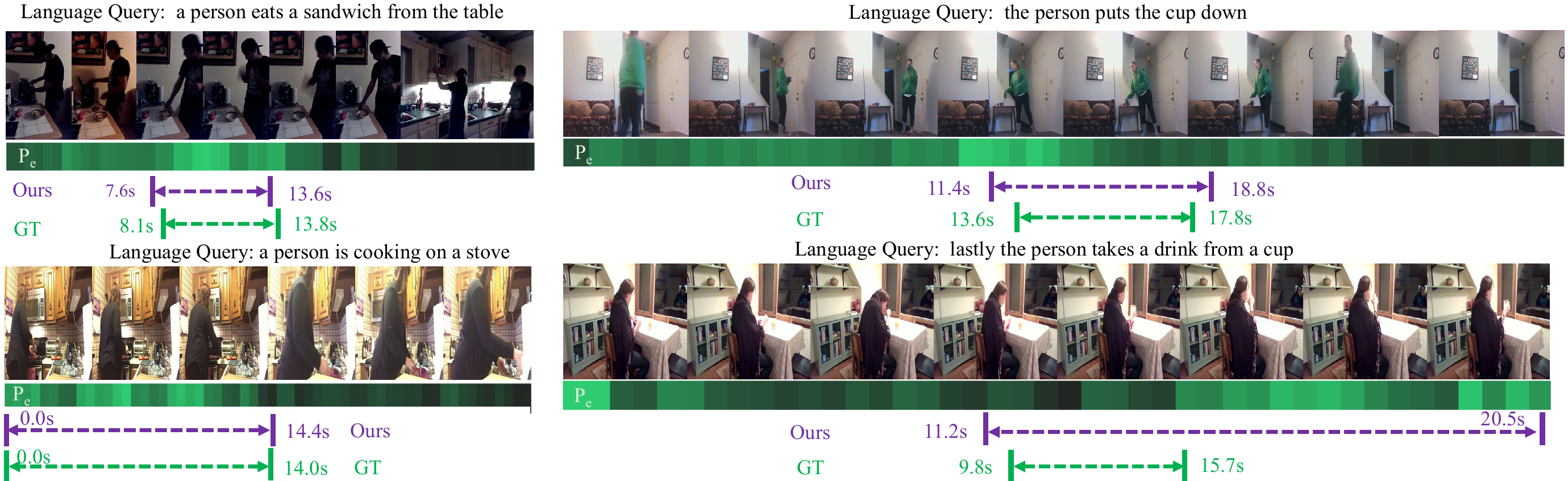}
    \caption{Examples of action localization visualization on the Charades-STA dataset. The action-relevant score $P_e$ is predicted by the entity-aware Transformer and brighter colors indicate higher values.}
    \label{fig:case}
\end{figure*}

\begin{figure}
    \centering
    \includegraphics[width=0.9\linewidth]{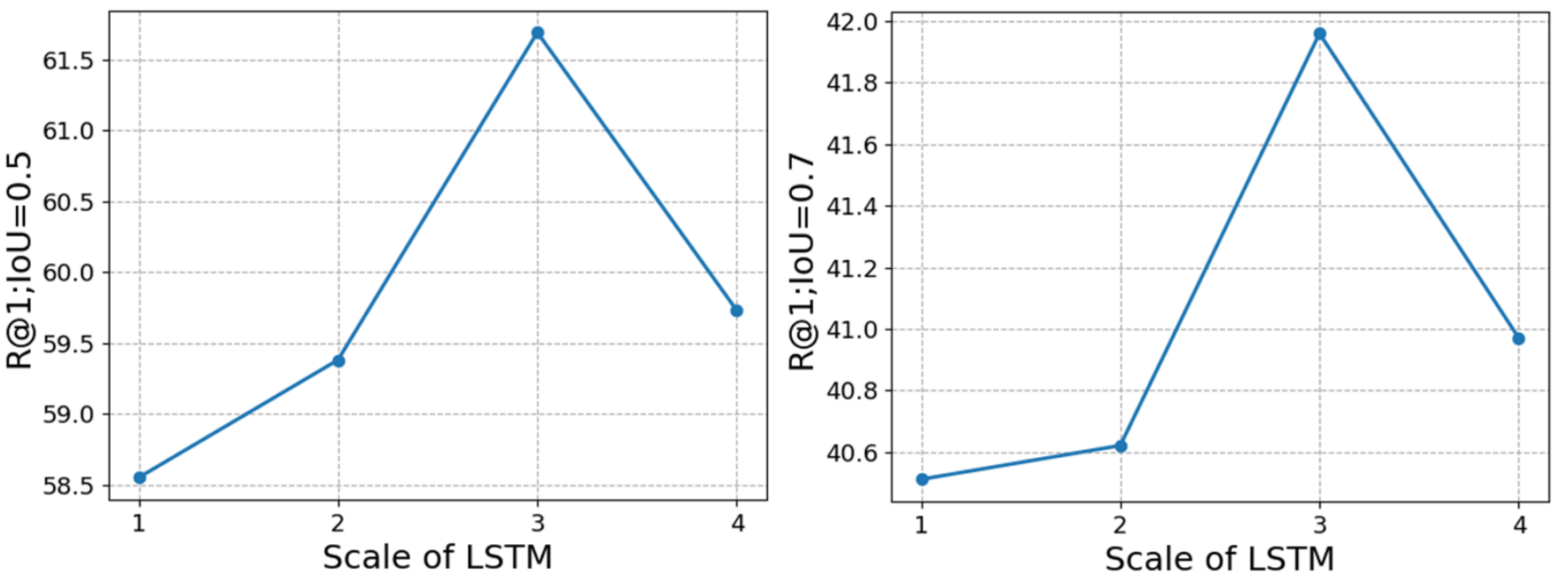}
    \caption{Analysis of the effect of scale number in LSTM Transformer on the Charades-STA dataset.}
    \label{fig:param_scales}
\end{figure}

\subsection{Ablation Studies}
We perform in-depth ablation studies to evaluate each component of our method on the Charades-STA dataset. The results are shown in Table \ref{tab:abl_mat}.

\noindent \textbf{Effect of Entity-aware Transformer.}
To evaluate the  entity-aware Transformer, we design a baseline model called ``Ours w/o EA Trans" that uses only the motion-aware Transformer with the input word feature of the language query.
As shown in Table \ref{tab:abl_mat}, our method outperforms ``Ours w/o EA Trans" with gains of 3\%  on all evaluation metrics, clearly demonstrating the effectiveness of the entity-aware Transformer. 

\begin{table}
\small
\centering
\begin{tabular}{l|ccc|c}
\hline
\hline
\multicolumn{1}{c|}{\multirow{2}{*}{Methods}}   & \multicolumn{3}{c|}{$R@1; IoU = \mu$} & \multirow{2}{*}{$mIoU$} \\
            & \multicolumn{1}{c}{0.3}  & \multicolumn{1}{c}{0.5}  & \multicolumn{1}{c|}{0.7} \\ \hline
BPNet~\cite{xiao2021boundary}  & 25.96 & 20.96 & 14.08 & 19.53 \\
VSLNet~\cite{zhang2021natural}    & 29.61 & 24.27 & 20.03 & 24.11 \\
I$^2$N~\cite{ning2021interaction}  & 31.80   & 28.69 & - & - \\
SS~\cite{ding2021support}     & 41.33   & 29.56 & - & -\\
CPNet~\cite{li2021proposal}  & 42.61   & 28.29 & - & 28.69 \\
CBLN~\cite{liu2021context}   & 38.89 & 27.65 & - & - \\ 
ICG~\cite{nan2021interventional}    & 38.84 & 29.07 & 19.05 & 28.26 \\
SMIN~\cite{wang2021structured}   & 48.01 &  35.24 & - & - \\
CPN~\cite{zhao2021cascaded}    & 48.29 & 36.58 & 21.25 & 34.63 \\
\hline
\multicolumn{1}{c|}{Ours}    & \textbf{50.11} & \textbf{38.16} & \textbf{26.82} & \textbf{36.43} \\
\hline \hline			
\end{tabular}
\caption{Comparison with the state-of-the-art methods on the TACoS dataset.}
\label{tab:sota_tacos}
\end{table}

\begin{table}
\small
\centering
\begin{tabular}{l|ccc|c}
\hline
\hline
\multicolumn{1}{c|}{\multirow{2}{*}{Methods}}   & \multicolumn{3}{c|}{$R@1; IoU = \mu$} & \multirow{2}{*}{$mIoU$} \\
            & \multicolumn{1}{c}{0.3}  & \multicolumn{1}{c}{0.5}  & \multicolumn{1}{c|}{0.7} \\ \hline
Ours w/o EA Trans             & 71.15 & 58.25 & 38.79 & 51.87 \\ \hline
FC Trans  &  67.18 & 48.14 & 27.69 & 46.24 \\		
T-Conv Trans  &  73.60 & 55.86 & 37.39 & 53.20 \\	 	\hline	
T-Conv  & 61.64 & 37.34 & 21.91 & 42.50 \\
LSTM    & 72.34 & 59.14 & 40.53 & 53.12 \\ \hline
Ours  &  \textbf{74.19} & \textbf{61.69} & \textbf{41.96} & \textbf{54.45} \\	
\hline \hline
\end{tabular}
\caption{Ablation studies on the Charades-STA dataset. %
}
\label{tab:abl_mat}
\end{table}

\noindent \textbf{Analysis of Motion-aware Transformer.}
To evaluate the Motion-aware Transformer, we design several variants of our method for comparison, denoted as ``FC Trans", ``T-Conv Trans",  ``T-Conv" and ``LSTM":
(i) ``FC Trans" and ``T-Conv Trans" replace the LSTM cell by fully connected layers and temporal convolutional layers, respectively. So ``FC Trans" degrades into a standard Transformer;
(ii) ``T-Conv" and ``LSTM" replace the LSTM Transformer by temporal convolutional and LSTM layers, respectively.
For a fair comparison, ``T-Conv Trans" and ``T-Conv" have multiple kernel sizes of 3, 5, and 7, and ``LSTM" has the same multi-scale LSTM as LSTM Transformer.
From the result in Table \ref{tab:abl_mat}, we have the following observations. 
(1) Compared with ``T-Conv Trans", our method achieves better results with gains of 5.83\% on $R@1; IoU = 0.5$ and 4.57\% on $R@1; IoU = 0.7$. Moreover, ``T-Conv Trans" outperforms ``FC Trans" by 7.72\% on $R@1; IoU = 0.5$ and 9.70\% on $R@1; IoU = 0.7$. These validate that carefully modeling of local motion changes significantly improves the localization accuracy.
(2) Compared with ``LSTM", our method achieves better results. %
Moreover, ``T-Conv Trans" outperforms ``T-Conv" by more than 10\% on all evaluation metrics. These show that the global changes captured by Transformer are beneficial to action localization.

\subsection{Parameter Analysis}
\noindent \textbf{Temporal Scale Number in LSTM Transformer.}
The performances of different temporal scales in LSTM Transformer on the Charades-STA dataset are  shown in Figure \ref{fig:param_scales}. We observe that when the scale number increases, the performance first increases and then gradually decreases, which demonstrates that modeling local motion changes at more temporal scales can improve the location precision, but also may bring redundant information.

\begin{figure}
    \centering
    \includegraphics[width=0.9\linewidth]{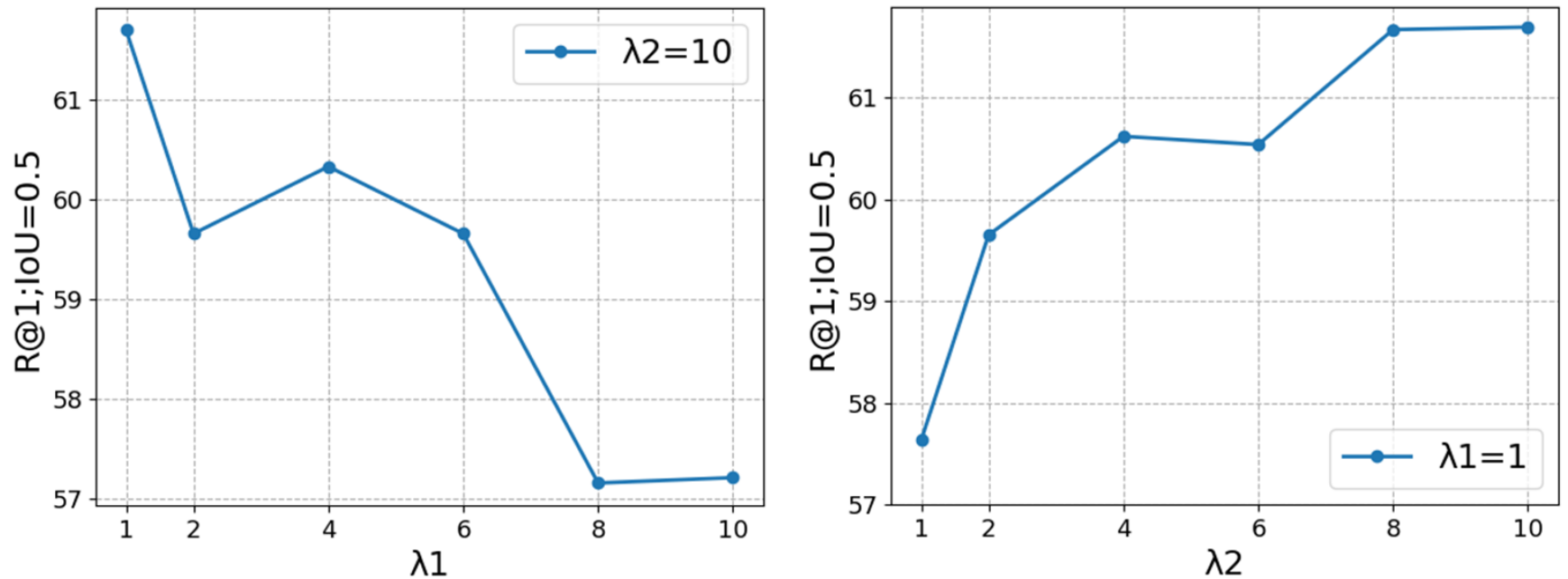}
    \caption{Analysis of the effect of loss weights on the Charades-STA dataset.}
    \label{fig:param_loss}
\end{figure}

\noindent \textbf{Loss Weights.}
To analyze the effect of the loss weights $\lambda_1$ and $\lambda_2$ in Equation (\ref{equ:loss}), we vary the value of $\lambda_1$ in [1,10] and the value of $\lambda_2$ in [1,10]. The results are shown in Figure \ref{fig:param_loss}. It is interesting to observe that when  $\lambda_1$ increases, the performance drops dramatically.
In contrast, the performance improves along with the increasing $\lambda_2$,
which shows that larger $\lambda_2$ boosts the per-frame inner prediction, and thus helps the boundary localization.

\subsection{Qualitative Analysis.}
We show several examples of action localization results on the Charades-STA dataset in Figure \ref{fig:case} by visualizing the corresponding action-relevant scores  predicted by the entity-aware Transformer where bright colors indicate higher values.
From the first three cases, we see that the action-relevant scores make it easier to accurately localize the action boundaries by paying more attention to the target action in a shrunken temporal.
However, in the last case, the entities (``\textit{person \& cup}") remain unchanged that the action-relevant scores of different frames are similar (the margin between maximum and minimum is less than 0.1), thus contributing less to the final boundary localization. 
 Moreover, the entity word ``\textit{drink}" is wrongly classified to a motion word, so that the predicted boundaries wrongly fall in the boundaries of ``\textit{drink}".

\section{Conclusion}

We have presented a novel coarse-to-fine model called entity-aware and motion-aware Transformers for language-driven action localization. It can progressively predict the action boundaries with high precision by first attending the action-relevant clips via the entity-aware Transformer and then refining the start and end frames via the motion-aware Transformer. 
By integrating multi-scale long short-term memory cells into the self-attention module, the motion-aware Transformer succeeds in capturing the fine-grained motion changes, thus achieving promising results. Extensive experiments on two public datasets have demonstrated that our method outperforms the state-of-the-art methods. 
In the future work, we are going to apply the entity-aware and motion-aware Transformers to weakly-supervised language-driven action localization. 

\section*{Acknowledgments}
This work was supported in part by the Natural Science Foundation of China (NSFC) under Grant No 62072041.
\newpage
\bibliographystyle{named}
\bibliography{ijcai22.bib}
\end{document}